\documentclass[11pt]{article}

\usepackage[final]{acl}

\usepackage{times}
\usepackage{latexsym}
\usepackage[T1]{fontenc}
\usepackage[utf8]{inputenc}
\usepackage{microtype}
\usepackage{inconsolata}
\usepackage{graphicx}
\usepackage{subcaption}
\usepackage{booktabs}
\usepackage{enumitem}
\usepackage{multirow}
\usepackage{amsmath}
\usepackage{amssymb}
\usepackage{mathtools}
\usepackage{amsthm}
\usepackage[capitalize,noabbrev]{cleveref}

\theoremstyle{plain}

\theoremstyle{definition}

\theoremstyle{remark}

\setlist[itemize]{leftmargin=*,topsep=2pt,itemsep=1pt}
\setlist[enumerate]{leftmargin=*,topsep=2pt,itemsep=1pt}
\AtBeginDocument{%
  \setlength{\abovedisplayskip}{4pt plus 1pt minus 2pt}%
  \setlength{\belowdisplayskip}{4pt plus 1pt minus 2pt}%
  \setlength{\abovedisplayshortskip}{2pt plus 1pt minus 1pt}%
  \setlength{\belowdisplayshortskip}{2pt plus 1pt minus 1pt}%
}

\title{Budget-Aware Compression Pipeline for Single-GPU LLM Inference: Methods, Trade-offs, and Coupling Effects}

\author{Hongyu Yu\thanks{~~Equal Contributions. Correspondence: Yifei Shen (\texttt{yifeis02@uw.edu})}~$^{1}$ \quad Yifei Shen$^{*}$~$^{2}$ \vspace{2pt}\\ $^{1}$ Lenovo Research \quad $^{2}$ University of Washington}

\begin{document}
\maketitle

\begin{abstract}
Single-GPU deployment of 70B-parameter language models on an NVIDIA GPU is constrained by device memory, long-context throughput, and engineering integration cost. We cast single-GPU inference as a budget-aware design problem over these three axes and study how pruning, quantization, and KV-cache compression interact under realistic execution. Controlled ablations show that layer-wise pruning makes weight quantization more robust. KV-cache sparsification complements INT8 KV quantization by reducing memory without hurting decoding speed, while static vector quantizers often conflict with dynamic caching. Guided by these coupling results and explicit budget tracking, we assembled a practical pipeline and compressed a 70B model to about 33 GB, sustained about 57 tokens/s on 10k token prompts on a single A40, and kept absolute accuracy within 5\% on common and reasoning benchmarks. We contribute design rules and a reproducible evaluation protocol that jointly report quality, memory, and end-to-end speed, and we provide a foundation for automated pipeline search under realistic single-GPU constraints.
\end{abstract}

\section{Introduction}
\label{sec:intro}

Large language models (LLMs) have grown rapidly in parameter count and context length, pushing the memory and compute requirements of inference beyond what a single commodity GPU can provide in straightforward FP16 deployments. For instance, a 70B-parameter model in half precision requires well over 100~GB of device memory, while widely available accelerators such as the NVIDIA A40 offer only 48~GB of VRAM. Long-context inference further amplifies the footprint due to the KV-cache that scales linearly with sequence length. As a result, deploying state-of-the-art LLMs on a single GPU is often bottlenecked simultaneously by (i) memory, (ii) compute/latency, and (iii) integration and compiler constraints \citep{zhou2024efficient}.

A large body of work proposes point solutions in three directions: \emph{post-training quantization} (PTQ) to reduce weight/activation precision \citep{lin2023awq,frantar2022gptq,xiao2022smoothquant,ashkboos2024quarot,tseng2024qtip,tseng2024quipsharp,lin2024duquant,shao2023omniquant}, \emph{pruning and sparsification} to eliminate parameters and FLOPs \citep{sun2023wanda,ashkboos2024slicegpt,men2024shortgpt}, and \emph{KV-cache compression} to tame the run-time memory of long sequences \citep{cai2024pyramidkv,liu2024kivi,hooper2024kvquant,frantar2023sparsegpt}. However, the community still lacks a \emph{unified}, budget-aware pipeline that combines these techniques under realistic single-GPU constraints. In practice, the best standalone method rarely remains best under composition: stacking introduces coupling that can be beneficial or harmful. Production outcomes are further constrained by integration details such as \emph{torch.compile}'s static-shape assumptions and engine-specific KV-cache layouts.
\begin{figure}[t]
\centering
\includegraphics[width=0.72\linewidth]{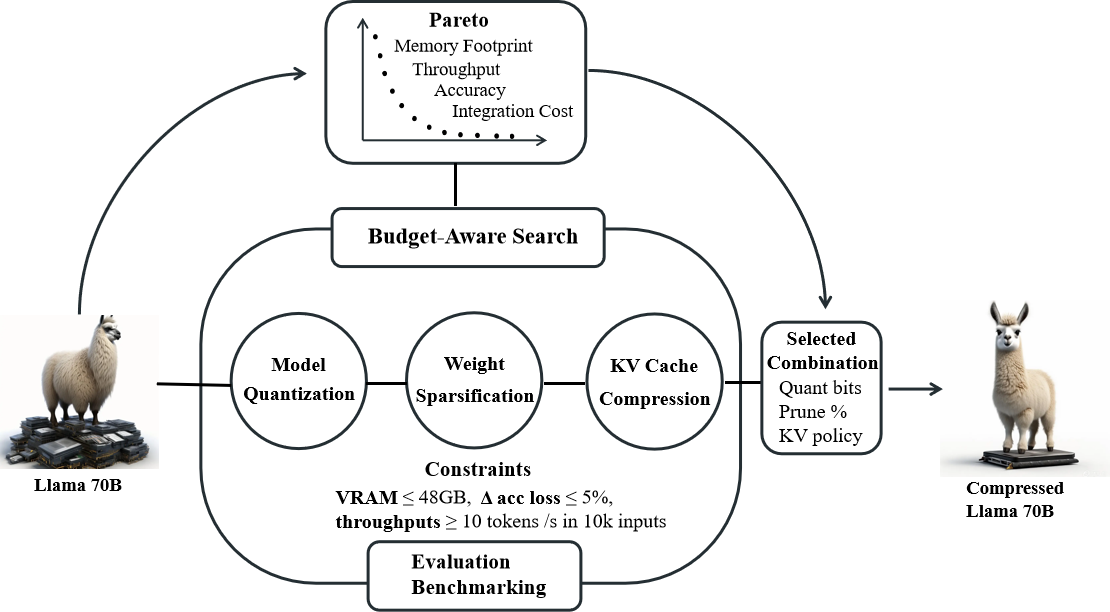}
\caption{Budget-aware single-GPU inference workflow. The three budgets (memory footprint, throughput/latency, integration effort) shape module selection across quantization, pruning, and KV-cache optimization.}
\label{fig:overview}
\end{figure}

\textbf{Goal and approach.} Our objective is to enable efficient inference of 70B-parameter language models on a \emph{single} NVIDIA A40 (48 GB) for long-context inputs ($\geq$10{,}000 tokens) under strict quality and latency constraints. We cast the design as an optimization under a three-dimensional budget: memory footprint, throughput/latency, and integration cost, and develop a unified compression pipeline that integrates quantization, pruning, and KV-cache sparsification. Concretely, our implementation applies 4-bit weight-only quantization, followed by layer-wise weight pruning, and a hierarchical KV-cache sparsification scheme combined with 8-bit KV cache quantization. This configuration fits a 70B model together with its 10k-token KV cache within 48 GB of device memory and limits accuracy degradation to absolute $\leq 5\%$ on reasoning and commonsense benchmarks.

\paragraph{Contributions.} This work makes four contributions.
\begin{enumerate}
\item \textbf{Unified, budget-aware pipeline.} We formulate single-GPU LLM inference as a memory, latency, and integration joint budgeting problem and present a practical pipeline that combines PTQ, structured pruning, and KV cache compression into a single workflow that runs on commodity hardware.
\item \textbf{Design principles under coupling.} We characterize positive/negative interactions among techniques and distill actionable rules that avoid common failure modes while preserving accuracy. Each rule is supported by targeted ablations isolating the mechanism responsible for the observed gain or degradation.
\item \textbf{End-to-end validation.} On a 70B model, the pipeline achieves fit within 48~GB, supports $\geq$10k-token inputs, and meets $\geq$10~tokens/s with $\leq$5\% absolute accuracy loss, demonstrating practical single-GPU deployment. 
\item \textbf{Toward automated pipeline search.} We outline an extensible search space over quantization bit widths, pruning ratios, and KV retention rates, allowing Pareto exploration under explicit budgets.
\end{enumerate}

Taken together, our results indicate that \emph{composing} lightweight compression techniques, rather than optimizing any one in isolation, is key to practical single-GPU inference of modern LLMs.

\begin{figure*}[t]
\centering
\includegraphics[width=0.62\linewidth]{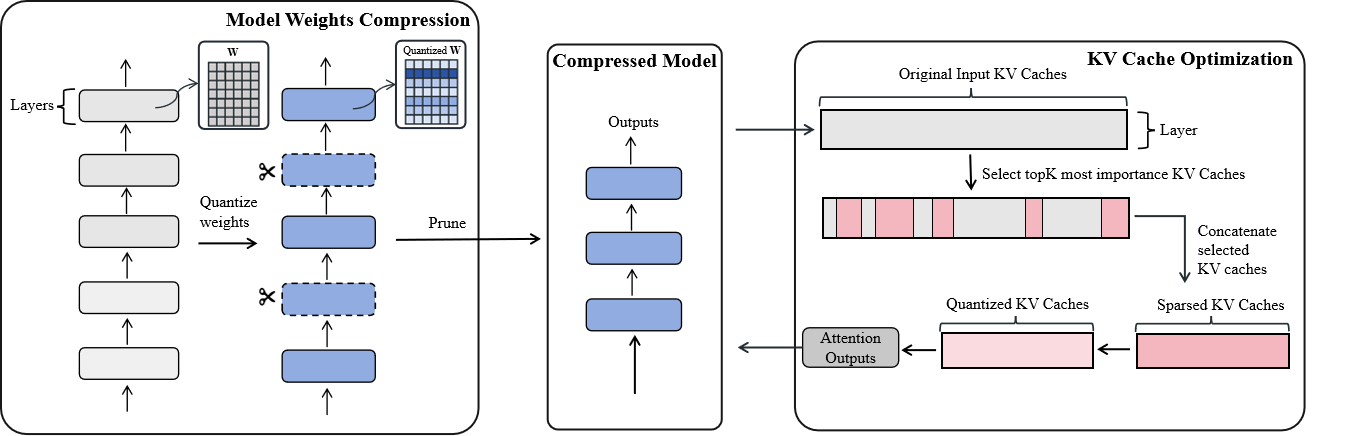}
\caption{Unified AWQ\,$\rightarrow$\,ShortGPT\,$\rightarrow$\,PyramidKV pipeline. AWQ delivers 4-bit weights with minimal loss, structured pruning reduces compute and stabilizes activations and PyramidKV plus INT8 cache quantization keeps long-context KV memory within the 48\,GB A40 budget.}
\label{fig:pipeline}
\end{figure*}

\section{Related Work}
\label{sec:related}

\paragraph{Post-training Quantization}
Post-training quantization (PTQ) is the most direct way to shrink LLM weights for single-GPU inference. Weight-only PTQs such as AWQ and GPTQ preserve accuracy at 4-bit by protecting activation-sensitive channels (AWQ) or solving a local second-order approximation (GPTQ) \citep{lin2023awq,frantar2022gptq}. Going beyond weight-only, SmoothQuant reduces activation outliers to enable W\#A\# configurations with small loss \citep{xiao2022smoothquant}. Rotation-based methods (e.g., QuaRot) mitigate outliers via orthogonal transforms for stable 4-bit inference \citep{ashkboos2024quarot}. Vector quantization families (e.g. QuIP \# and QTIP) push to very low bit rates (down to 2-bit) by codebooks/trellises while maintaining quality, although at higher integration cost in current engines \citep{tseng2024quipsharp,tseng2024qtip}. Outlier distribution approaches such as DuQuant and omnidirectional calibration (OmniQuant) further improve robustness at low precision \citep{lin2024duquant,shao2023omniquant}.

\paragraph{Pruning and Sparsity}
Structured and unstructured pruning reduce FLOPs/VRAM and can complement quantization. Unstructured pruning constructs sparse weight matrices by ranking parameter importance and setting low-importance entries to zero. SparseGPT \citep{frantar2023sparsegpt} estimates importance using squared weights scaled by an approximation to the inverse Hessian, whereas Wanda \citep{sun2023wanda} relies on the elementwise product of layer weights and activation magnitudes. 

Structure-altering methods such as SliceGPT operate at the width level by removing rows and columns that are provably redundant under an orthogonal reparameterization of the weight matrices \citep{ashkboos2024slicegpt}. Layer pruning as in ShortGPT assumes that the network depth is over-parameterized and scores transformer blocks using BI coefficients, removing those ranked as least important \citep{men2024shortgpt}. 


\paragraph{KV-Cache Compression and Long-Context Optimization}
The KV cache is often the dominant runtime memory for long contexts. Approaches fall into three buckets: (i) quantization of KV (e.g., KIVI, KVQuant) to 2--3 bits with minimal perplexity loss \citep{liu2024kivi,hooper2024kvquant}; (ii) eviction/streaming policies (StreamingLLM; H$_2$O) that drop less-useful tokens dynamically to bound memory \citep{zhang2023h2o}; and (iii) sparse retrieval/compression that keeps only salient KV entries (SnapKV, ShadowKV) or allocates budget adaptively across layers (PyramidKV) \citep{li2024snapkv,sun2025shadowkvkvcacheshadows,cai2024pyramidkv}. 

These methods substantially shrink per-sequence memory and enable longer contexts on a single GPU, but differ in throughput/memory trade-offs and engine compatibility.

\paragraph{Serving Systems, Kernels, and Compiler Constraints}
Modern serving systems co-design memory and scheduling to maximize single-GPU throughput. vLLM introduced PagedAttention, which eliminates KV-cache fragmentation with a paged layout and enables continuous batching and cache sharing; FlashAttention/FlashAttention-2 provide exact, IO-aware attention kernels that many engines integrate; SGLang adds program-level optimizations (e.g., radix/prefix caching) for structured LM workloads \citep{kwon2023vllm,dao2022flashattention,zheng2023sglang}. In practice, compiler/toolchain constraints (e.g., static shape requirements and kernel fusion limits) also influence which compression combinations are feasible in production engines, an issue reflected in our system integration.

Despite rapid advances in quantization, pruning, and KV-cache management, prior work typically evaluates each technique in isolation. As a result, the field lacks unified experimental designs that measure interaction effects across techniques. For instance, how pruning alters activation/weight outlier distributions that quantization must handle, or how KV eviction/quantization reshapes attention dynamics and thus the efficacy of sparsity patterns. Existing surveys primarily catalog method families but rarely offer controlled factorial ablations across method combinations \cite{zhou2024efficient,zhu2024surveymodelcompressionlarge,li2025surveylargelanguagemodel}. We address this gap through a systematic comparative study of compression methods, quantifying their synergies and interference under a unified evaluation protocol. The resulting evidence identifies robust, mutually compatible combinations that form a principled and composable recipe for single-GPU inference.

\section{Problem Formulation and Budget-Aware Analysis Framework}  
\label{sec:problem}

Deploying 70B-parameter language models on a single NVIDIA A40 (48 GB) poses fundamental challenges due to the large memory footprint of both model weights and runtime activations. The FP16 parameters alone exceed 140 GB, while the KV-cache grows linearly with the input sequence length \citep{dao2022flashattention, kwon2023vllm}. To make such large-scale models feasible for long-context inference (e.g., $\geq$10{,}000 tokens) within limited hardware budgets, it is essential to compress the model while maintaining generation quality and throughput. We explore a spectrum of post-training compression strategies, including quantization, structured pruning, and KV-cache sparsification, and identify an effective combination that meets strict memory and latency constraints. Specifically, we formulate the deployment problem as satisfying a three-dimensional budget encompassing (1) memory footprint, (2) throughput/latency, and (3) integration cost, inspired by recent budget-aware optimization frameworks for efficient LLM inference \citep{zhou2024efficient, wan2024efficientLLM}.

\subsection{Deployment Budget and Constraints}
The deployment budget is formalized as $\mathcal{B} = (B_{\mathrm{mem}}, B_{\mathrm{thr}}, B_{\mathrm{int}})$.
The memory budget $B_{\mathrm{mem}}$ caps the combined footprint of weights, activations, and KV caches below the 48\,GB capacity of the A40.
The end-to-end throughput budget $B_{\mathrm{thr}}$ mandates at least 10 tokens/sec under 10k-token contexts, reflecting our production latency envelope.
The integration budget $B_{\mathrm{int}}$ quantifies the engineering cost of adopting a method on an existing serving stack (kernel availability, compiler assumptions, and compatibility with dynamic batching).
Model quality is treated as a strict constraint: downstream accuracy must remain within a 5\% margin on common and reasoning task evaluations, such as Arc-easy \citep{allenai:arc} and GPQA-Diamond \citep{rein2024gpqa}.

\subsection{Compression Modules and Decision Variables}
To satisfy $\mathcal{B}$ we compose a pipeline from three module families whose hyperparameters define the decision space.
\textbf{Post-training quantization (PTQ)} covers weight-only methods (AWQ, GPTQ), weight-activation techniques (SmoothQuant, QuaRot), and vector/rotation-based approaches (QuIP\#, QTIP, DuQuant, OmniQuant) \citep{lin2023awq,frantar2022gptq,xiao2022smoothquant,ashkboos2024quarot,tseng2024quipsharp,tseng2024qtip,lin2024duquant,shao2023omniquant}.
\textbf{Pruning and sparsification} spans layer removal (ShortGPT), width slimming (SliceGPT), and weight sparsity (Wanda, SparseGPT) \citep{men2024shortgpt,ashkboos2024slicegpt,sun2023wanda,frantar2023sparsegpt}.
\textbf{KV-cache optimization} includes retrieval-aware caching (SnapKV, H$_2$O, StreamingLLM), hierarchical eviction (PyramidKV), and quantized KV storage (KVQuant, KIVI) \citep{li2024snapkv,zhang2023h2o,cai2024pyramidkv,hooper2024kvquant,liu2024kivi,xiao2023streamingllm}.
These modules interact: pruning alters activation distributions that quantizers must absorb, KV eviction changes attention locality and thus sparsity effectiveness, and vector quantizers impose static-shape constraints that can clash with dynamic KV sparsity. Our framework therefore evaluates both standalone configurations and composite pipelines.

Figure~\ref{fig:pipeline} summarizes the pipeline geometry and highlights where each module consumes the memory, throughput, and integration budgets. The schematic also anticipates the ordering and compatibility constraints we explore in later sections.

\subsection{Budget-Aware Evaluation Protocol}
Each pipeline is instantiated on DeepSeek-R1-Distill-Llama-70B checkpoints \citep{deepseekai2025deepseekr1}. Measurements capture (i) peak device memory via NVML profiling during 10k-token decoding, (ii) generated tokens per second under Transformers framework \citep{wolf-etal-2020-transformers}, and (iii) integration effort, scored on a three-level scale capturing required code modifications (plug-and-play, kernel adaptation, or compiler reimplementation).
For quality, we track perplexity on WikiText2/C4 \citep{merity2016pointer, raffel2020exploring} for screening and task accuracy on QA and commonsense subsets aligned with our target applications.
The framework also records ordering effects, such as applying ShortGPT before AutoAWQ versus post-quantization pruning, to surface positive coupling and negative coupling.

\section{Single-Method Effects under Single-GPU Budgets}
\label{sec:single}

We begin with controlled ablations of each compression family on LLaMA to quantify their isolated contributions under the deployment budget from \Cref{sec:problem}.

\subsection{Post-Training Quantization}
\label{sec:ptq}

\begin{table*}[t]
\centering
\scriptsize
\caption{Perplexity screening of post-training quantization on 8B-class checkpoints under single-A40 constraints. Lower perplexity is better.}
\label{tab:quant_screen}
\setlength{\tabcolsep}{4pt}
\renewcommand{\arraystretch}{0.86}
\begin{tabular}{@{}l l l c c c@{}}
\toprule
Method & Category & Model & Bits & WikiText2 & C4 \\
\midrule
SmoothQuant & Activation-aware & LLaMA~3.1~8B & W4A4 & 220.16 & 196.83 \\
QuaRot & Activation-aware & LLaMA~2~7B & W4A4 & 11.86 & N/A \\
QuaRot & Activation-aware & LLaMA~3.1~8B & W4A4 & 17.19 & N/A \\
QuaRot + GPTQ & Activation-aware & LLaMA~2~7B & W4A4 & 8.32 & N/A \\
QuaRot + GPTQ & Activation-aware & LLaMA~3.1~8B & W4A4 & 13.38 & N/A \\
GPTQ (AutoGPTQ) & Weight-only & LLaMA~3.1~8B & 4-bit & 7.00 & 11.80 \\
AWQ (AutoAWQ) & Weight-only & LLaMA~3.1~8B & W4A16 & 6.77 & N/A \\
QTIP & Vector quantization & LLaMA~3~8B & 4-bit & 5.67 & 7.20 \\
QTIP & Vector quantization & LLaMA~3~8B & 2-bit & 7.33 & 8.62 \\
QuIP\# & Vector quantization & LLaMA~3~8B Instruct & 2-bit & 11.23 & 14.71 \\
DuQuant & Vector quantization & LLaMA~3.1~8B & W4A4 & 9.95 & 13.69 \\
OmniQuant & Specialized & LLaMA~3.1~8B Instruct & W4A16 & 8.97 & N/A \\
\bottomrule
\end{tabular}
\end{table*}

Table~\ref{tab:quant_screen} shows that low-bit PTQ substantially reduces perplexity only when the method controls activation outliers. SmoothQuant increases WikiText2 perplexity to 220.16 even at W4A4, while QuaRot remains competitive on LLaMA~2~7B (11.86) but degrades to 17.19 on LLaMA~3.1~8B. Coupling QuaRot with GPTQ lowers the 7B score to 8.32 but still lags on 8B at 13.38. AWQ and GPTQ deliver the strongest weight-only results across both datasets, and vector quantizers such as QTIP and QuIP\# push to 2-bit regimes but incur higher integration overhead that complicates coupling with pruning and KV sparsity \citep{lin2023awq,frantar2022gptq,tseng2024qtip,tseng2024quipsharp}. DuQuant provides moderate perplexity yet demands additional activation calibration, keeping it outside the budget that prioritizes low latency.

\begin{table*}[t]
\centering
\scriptsize
\caption{INT4 post-training quantization on Distilled Llama 3 70B evaluated on a single NVIDIA A40. Accuracy metrics are higher-is-better.}
\label{tab:quant_70b}
\setlength{\tabcolsep}{4pt}
\renewcommand{\arraystretch}{0.86}
\begin{tabular}{@{}l c c c c c c c@{}}
\toprule
Method & Model Size (GB) & WikiText2 $\downarrow$ & ARC-C $\uparrow$ & ARC-E $\uparrow$ & Winogrande $\uparrow$ & AIME-24 $\uparrow$ & GPQA-D $\uparrow$ \\
\midrule
AWQ (INT4) & 38 & 6.563 & 0.546 $\pm$ 0.014 & 0.809 $\pm$ 0.008 & 0.742 $\pm$ 0.013 & 0.400 & 0.636 \\
GPTQ (INT4) & 38 & 5.743 & 0.510 $\pm$ 0.015 & 0.802 $\pm$ 0.008 & 0.732 $\pm$ 0.013 & 0.467 & 0.561 \\
\bottomrule
\end{tabular}
\end{table*}

Table~\ref{tab:quant_70b} confirms AWQ as the most balanced INT4 option on DeepSeek-R1-Distill-Llama-70B. Although GPTQ attains lower WikiText2 perplexity, AWQ achieves higher accuracy on ARC-Challenge, ARC-Easy, Winogrande, and GPQA \citep{allenai:arc, sakaguchi2021winogrande, rein2024gpqa} with identical 38\,GB artifacts and integrates cleanly with mainstream engines \citep{lin2023awq}. These observations motivate using AWQ as the baseline quantizer before introducing additional compression stages.

\subsection{Pruning and Sparsification}
\label{sec:pruning}

\begin{table}[t]
\centering
\scriptsize
\caption{Wanda pruning on LLaMA~3.1~8B across sparsity levels. Accuracy metrics are mean $\pm$ standard error.}
\label{tab:wanda}
\setlength{\tabcolsep}{3pt}
\renewcommand{\arraystretch}{0.86}
\begin{tabular}{@{}l c c c c@{}}
\toprule
Metric & Dense (0\%) & 25\% & 50\% & 75\% \\
\midrule
Model size (GB) & 14.96 & 14.96 & 14.96 & 14.96 \\
WikiText2 PPL $\downarrow$ & 8.29 & 8.39 & 11.97 & 229.82 \\
ARC-Challenge $\uparrow$ & 0.5299 & 0.5137 & 0.4172 & 0.1809 \\
ARC-Easy $\uparrow$ & 0.8152 & 0.8068 & 0.7441 & 0.3018 \\
Throughput (tok/s) $\uparrow$ & 34.40 & 35.70 & 34.75 & 24.23 \\
\bottomrule
\end{tabular}
\end{table}

\begin{table*}[t]
\centering
\scriptsize
\caption{SliceGPT pruning on LLaMA~3.1~8B Benchmark}
\label{tab:slicegpt}
\setlength{\tabcolsep}{4pt}
\renewcommand{\arraystretch}{0.86}
\begin{tabular}{@{}l c c c c c@{}}
\toprule
\multirow{2}{*}{Width setting} & \multirow{2}{*}{Model size (GB)} & \multirow{2}{*}{WikiText2 (PPL$\downarrow$)} & \multirow{2}{*}{PIQA (Acc$\uparrow$)} & \multicolumn{2}{c}{Throughput (tok/s)} \\
\cmidrule(lr){5-6}
 &  &  &  & Input 2000, Output 256 & Input 4000, Output 256 \\
\midrule
Dense (100\%) & 14.96 & 8.28  & 0.786 & 12.55 & 9.49 \\
75\% width    & 13.71 & 23.74 & 0.609 & 27.67 & 23.87 \\
50\% width    & 10.71 & 99.76 & 0.518 & 34.37 & 29.18 \\
\bottomrule
\end{tabular}
\end{table*}

\begin{table}[t]
\centering
\scriptsize
\caption{ShortGPT layer pruning on DeepSeek-R1 Distilled Llama 70B, with and without AWQ quantization.}
\label{tab:shortgpt}
\setlength{\tabcolsep}{4pt}
\renewcommand{\arraystretch}{0.86}
\begin{tabular}{@{}l c c c@{}}
\toprule
Configuration $\uparrow$ & Memory (GB) $\downarrow$ & ARC-E $\uparrow$ & GPQA-D $\uparrow$ \\
\midrule
Dense & 132.04 & 0.8152 & 0.6111 \\
Short-5 & 123.76 & 0.8043 & 0.5962 \\
Short-10 & 115.79 & 0.7934 & 0.5505 \\
Short-15 & 107.82 & 0.7582 & 0.4957 \\
Short-20 & 99.85 & 0.7222 & 0.2980 \\
AWQ & 37.34 & 0.8085 & 0.6364 \\
AWQ-5 & 35.27 & 0.7997 & 0.6111 \\
AWQ-10 & 33.20 & 0.7845 & 0.6010 \\
AWQ-15 & 31.13 & 0.7576 & 0.4848 \\
AWQ-20 & 29.06 & 0.7205 & 0.2980 \\
\bottomrule
\end{tabular}
\end{table}

\begin{table*}[t]
\raggedright
\scriptsize
\caption{Impact of pruning strategies on DeepSeek-R1-Distill-Llama-8B under single A40 evaluation. }
\label{tab:pruning}
\setlength{\tabcolsep}{3pt}
\renewcommand{\arraystretch}{0.86}
\begin{tabular}{@{}p{0.12\linewidth} p{0.13\linewidth} p{0.16\linewidth} r p{0.24\linewidth} p{0.11\linewidth}@{}}
\toprule
Method & Type & Portion & \multicolumn{1}{c}{$\Delta$ Thpt (\%)} & Accuracy (key deltas) & Tags \\
\midrule
Wanda
& Unstructured
& 0.5 sparsity
& +1.03
& \begin{tabular}[t]{@{}l@{}}PPL \(+44.57\%\)\\ ARC-C \(-20.75\%\)\end{tabular}
& [C1] \\

SliceGPT
& Width pruning
& 25\% width
& +151.53
& \begin{tabular}[t]{@{}l@{}}PPL \(+181.71\%\)\\ PIQA \(-22.78\%\)\end{tabular}
& [C2] \\

ShortGPT
& Layer pruning
& 3/32~layers
& +6.56
& \begin{tabular}[t]{@{}l@{}}PPL \(+35.01\%\)\\ ARC-C \(-4.10\%\)\end{tabular}
& [C3] \\
\bottomrule
\end{tabular}

\vspace{3pt}
\begin{minipage}{\linewidth}
\scriptsize
\textbf{Tag legend.} [C1] Conflicts with AWQ rescaling; reduces reliability. \quad
[C2] Requires kernel re-engineering; difficult to combine with low-bit kernels. \quad
[C3] Preserves architecture; integrates smoothly with AWQ.
\end{minipage}
\end{table*}

\begin{figure}[t]
  \centering
  \includegraphics[width=0.62\linewidth]{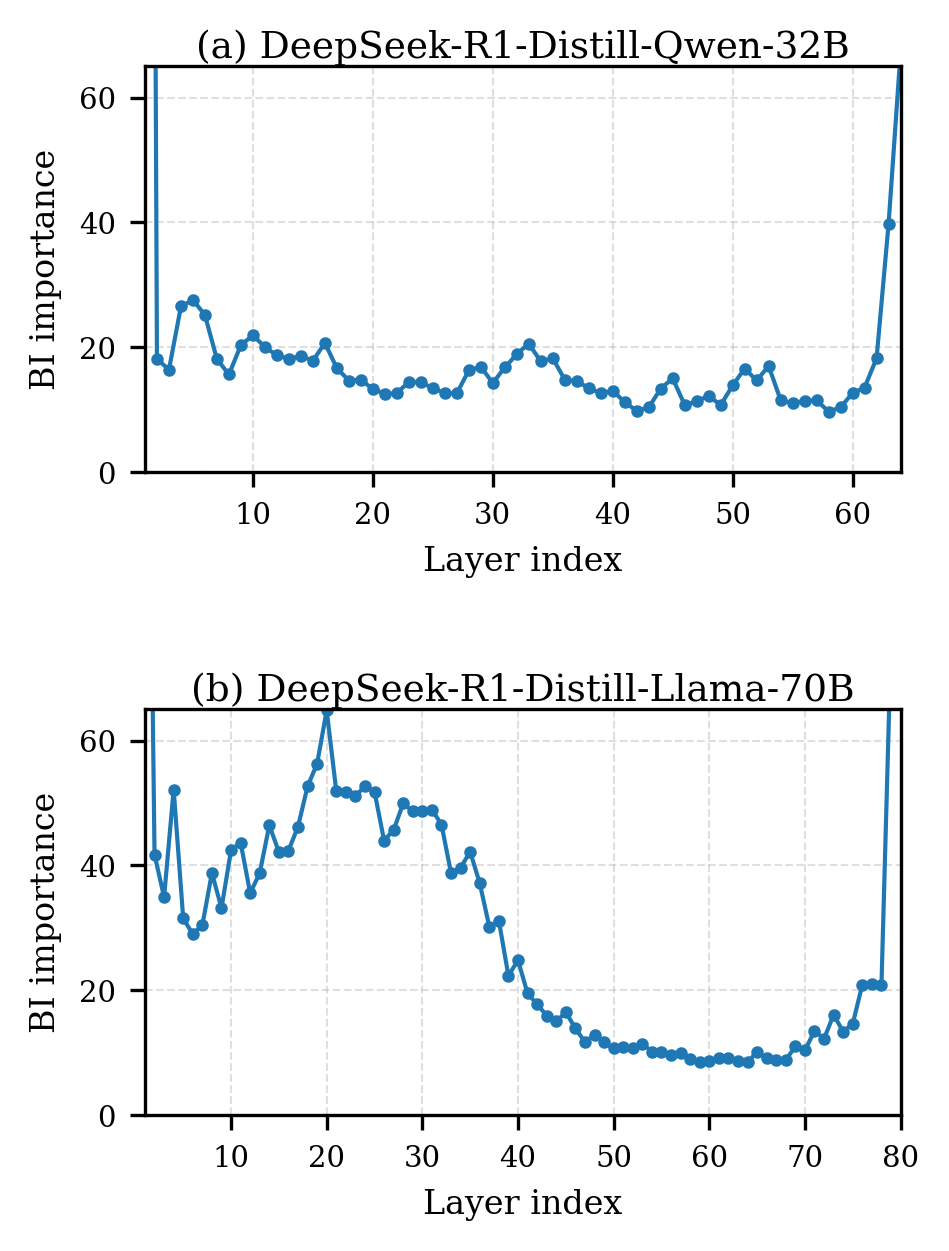}
  \caption{Layer-wise BI importance distributions for (a)  DeepSeek-R1-Distill-Qwen-32B and (b) DeepSeek-R1-Distill-Llama-70B.}
  \label{fig:bi-comparison}
\end{figure}

\subsubsection{Empirical limitations of depth-only ShortGPT pruning}
Table~\ref{tab:pruning} shows that Wanda yields negligible throughput gains and destabilizes activation quantization, consistent with AWQ's compatibility cautions. SliceGPT is more aggressive but degrades quality sharply at 25\% (perplexity $+181.71\%$, PIQA $-22.78\%$) and incurs substantial kernel re-engineering overhead. In contrast, ShortGPT offers the best trade-off at small ratios: pruning 3/32 layers \emph{after} AWQ improves throughput by 6.56\% over the dense baseline with only a 4.1\% Arc-challenge drop, while preserving structured topology and low integration cost.

A more detailed experiment is applied on DeepSeek-R1-Distill-Llama-70B model, and at higher pruning levels, ShortGPT's depth-only strategy saturates (Table~\ref{tab:shortgpt}): increasing from 10 to 20 pruned layers (25\% depth) causes steep losses on reasoning and knowledge tasks. A closer analysis of BI-based layer importance between two models, DeepSeek-R1-Distill-Llama-70B and DeepSeek-R1-Distill-Qwen-32B, reveals that this saturation is highly architecture dependent. In Fig.~\ref{fig:bi-comparison}, DeepSeek-R1-Distill-Llama-70B model exhibits smooth, banded importance profiles along depth, whereas DeepSeek-R1-Distill-Qwen-32B displays pronounced peaks in the middle layers. Therefore, pruning the model based on layer-wise importance coefficients alone may not be optimal.

These results motivate a more flexible, deployment-oriented pruning design that adapts to architecture-specific BI landscapes, which we present next.

\subsubsection{Continuity-aware layer reordering and planning.}

We first profile per-block importance using the BI coefficient~\citep{men2024shortgpt}.
Given hidden states $X_{i,t}$ and $X_{i+1,t}$ at consecutive blocks $i$ and $i{+}1$ for token $t$, we define
\begingroup
\small
\begin{equation}
\mathrm{BI}_i =
1-\mathbb{E}_{X,t}\!\left[
\frac{X_{i,t}^{\top}X_{i+1,t}}
{\lVert X_{i,t}\rVert_2\,\lVert X_{i+1,t}\rVert_2}
\right],\quad i=0,\ldots,N{-}1,
\end{equation}
\endgroup
where larger $\mathrm{BI}_i$ indicates a larger representational change and hence higher importance.
We estimate $\{\mathrm{BI}_i\}$ on a small calibration set and reuse it across pruning budgets.

\paragraph{Local continuity.}
To capture depth-wise smoothness, we compute a local continuity coefficient over a window of odd size $w=2m{+}1$.
Let $\mathrm{BI}_{\max}=\max_{1\le j \le N-1}\mathrm{BI}_j$ and define the clipped window
$W_k=[L_k,R_k]$ with $L_k=\max(2,k-m)$ and $R_k=\min(N-1,k+m)$.
We measure average BI variation within $W_k$ as
\begin{equation}
\mathrm{CC}_{\text{local}}(k)
\;=\;
1-\frac{1}{|W_k|-1}\sum_{i=L_k}^{R_k}\frac{\left|\mathrm{BI}_i-\mathrm{BI}_{i-1}\right|}{\mathrm{BI}_{\max}},
\end{equation}
where $|W_k|=R_k-L_k+1$.
A window is treated as a candidate contiguous pruning block if
$\mathrm{CC}_{\text{local}}(k)>\tau_c$ and its mean importance
$\overline{\mathrm{BI}}_{\text{win}}<\beta\cdot\mathrm{median}(\mathrm{BI})$.

\paragraph{Candidate generation.}
Given a pruning budget of $K$ blocks, we form two candidate pools.
$C_{\text{single}}$ contains the lowest-importance $\alpha K$ single blocks ranked by $\mathrm{BI}_i$
(excluding blocks already covered by any candidate window), and
$C_{\text{block}}$ contains low-importance windows ranked by their mean $\overline{\mathrm{BI}}_{\text{win}}$.
Unless stated otherwise, we use $\alpha{=}1.5$, $w{=}3$, $\tau_c{=}0.8$, and $\beta{=}0.8$.

\paragraph{Continuity-aware greedy planning.}
We represent each candidate including single block or window as an interval $c=[l_c,r_c]$.
For two intervals $c$ and $s$, define their gap distance
$d(c,s)=\max\!\bigl(0,\,l_s-r_c,\,l_c-r_s\bigr)$, and
$d(c,S)=\min_{s\in S}d(c,s)$ for the current selected set $S$.
Each candidate receives a base score
$\mathrm{score}(c)=-\overline{\mathrm{BI}}(c)$, where $\overline{\mathrm{BI}}(c)$ is the mean BI over $[l_c,r_c]$.
We add a continuity bonus if $d(c,S)\le r$:
\[
\mathrm{score}(c)\leftarrow \mathrm{score}(c)+\lambda.
\]
Starting from $S=\emptyset$, we repeatedly select the highest-scoring candidate,
add its interval to $S$, and update bonuses until $|S|=K$.
We use $r\in\{1,2\}$ and $\lambda=0.1$ in all experiments.

\begin{table*}[t]
\centering
\scriptsize
\setlength{\tabcolsep}{3pt}
\renewcommand{\arraystretch}{0.86}
\resizebox{\textwidth}{!}{%
\begin{tabular}{l c l c c}
\hline
\textbf{Model} & \textbf{Pruning layers} & \textbf{Pruned layer number} & \textbf{ARC\_E} & \textbf{GPQA-D} \\
\hline
dense    & -- & -- &
0.8068 & 0.6313 \\
shortgpt & 15 & [46, 50, 51, 53, 54, 55, 57, 58, 59, 61, 62, 64, 65, 66, 67] &
0.7534 & 0.4848 \\
shortgpt & 20 & [46, 47, 49, 50, 51, 53, 54, 55, 56, 57, 58, 59, 61, 62, 64, 65, 66, 67, 68, 71] &
0.7012 & 0.2879 \\
Layer Reorder       & 15 & [54, 55, 56, 57, 58, 59, 60, 61, 62, 64, 65, 66, 67, 68, 69] &
0.7580 & 0.5303 \\
Layer Reorder       & 20 & [50, 51, 52, 53, 54, 55, 56, 57, 58, 59, 60, 61, 62, 63, 64, 65, 66, 67, 68, 69] &
0.7243 & 0.4091 \\
\hline
\end{tabular}
}
\caption{The results across ShortGPT and Layer Reordering Pruning for Llama3-70B-AWQ model.}
\label{tab:pruning_reorder}
\end{table*}

Based on this continuity-aware planner, we re-optimized the pruning schedules for deeper budgets on DeepSeek-R1-Distill-Llama-70B and benchmarked them against the original ShortGPT schedules. Table~\ref{tab:pruning_reorder} compares the dense model, ShortGPT with 15 and 20 pruned layers, and our layer-reordering variant at the same pruning ratios. The dense baseline achieves 0.8068 on ARC\_E and 0.6313 on GPQA-Diamond. When pruning 15 of 80 layers, ShortGPT degrades to 0.7534 / 0.4848, while our reordered schedule attains 0.7580 / 0.5303, slightly improving ARC\_E and recovering a substantial portion of the reasoning loss on GPQA-Diamond. At the more aggressive 25\% depth reduction (20 layers), the gap becomes more pronounced: ShortGPT drops to 0.7012 / 0.2879, whereas our layer-reordering plan reaches 0.7243 / 0.4091, retaining markedly better accuracy on both benchmarks.

Importantly, the pruning schedules produced by our planner preferentially remove layers that form contiguous low-importance bands along depth, rather than the scattered layer subsets selected by BI scores only. These results indicate that, once layer importance is reordered under a continuity-aware criterion instead of relying solely on per-layer BI ranking, depth-only pruning can more safely operate at more aggressive budgets than previously feasible.


\subsection{KV-Cache Compression}
\label{sec:kv}

\begin{table*}[t]
\centering
\scriptsize
\caption{KV-cache optimization screening on LLaMA 3.1 8B. All runs use 11\,427 input tokens. TTFT denotes time-to-first-token.}
\label{tab:kv_screen}
\setlength{\tabcolsep}{5pt}
\renewcommand{\arraystretch}{0.86}
\begin{tabular}{@{}l c c c c@{}}
\toprule
Method & Peak Memory (MB) & TTFT (s) & Throughput (tok/s) & PPL \\
\midrule
Full KV & 4\,615 & 2.2132 & 180.5 & 6.50 \\
H$^2$O KV \citep{zhang2023h2o} & 3\,987 & 2.4985 & 210.4 & 6.55 \\
PyramidKV \citep{cai2024pyramidkv} & 3\,821 & 2.3456 & 220.3 & 6.60 \\
ShadowKV \citep{sun2025shadowkvkvcacheshadows} & 3\,505 & 2.1123 & 400.1 & 6.62 \\
SnapKV \citep{li2024snapkv} & 3\,702 & 2.2146 & 230.01 & 6.58 \\
\bottomrule
\end{tabular}
\end{table*}

KV-Cache sparsification and quantization govern the feasibility of 10k-token prompts on a 48\,GB device. Table~\ref{tab:kv_screen} reports all screening runs on 11\,427-token prompts and shows that, on 8B models, ShadowKV delivers the highest throughput but conflicts with subsequent quantization, whereas PyramidKV and SnapKV provide consistent memory savings with moderate latency overhead \citep{sun2025shadowkvkvcacheshadows,cai2024pyramidkv,li2024snapkv}. These techniques also retain low perplexity, validating them as candidates for the final system.

\begin{table}[t]
\centering
\scriptsize
\caption{KV-cache sparsification and quantization on DeepSeek-R1-Distilled Llama 70B. All runs use 11\,427-token inputs. Peak memory is reported via NVML.}
\label{tab:kv_70b}
\setlength{\tabcolsep}{1pt}
\renewcommand{\arraystretch}{0.86}
\resizebox{\linewidth}{!}{%
\begin{tabular}{@{}l c c c@{}}
\toprule
Configuration & Peak Memory (MB) & TTFT (s) & Throughput (tok/s) \\
\midrule
Full KV (FP16) & 40\,690 & 15.1636 & 68.51 \\
Full KV (INT8) & 39\,615 & 21.0056 & 55.75 \\
Full KV (INT4) & 38\,715 & 21.0423 & 53.63 \\
SnapKV (FP16) & 39\,985 & 15.7410 & 68.16 \\
SnapKV (INT8) & 39\,209 & 20.5750 & 57.25 \\
SnapKV (INT4) & 38\,542 & 20.7586 & 55.90 \\
PyramidKV (FP16) & 39\,835 & 15.6812 & 68.18 \\
PyramidKV (INT8) & 39\,029 & 20.8406 & 57.21 \\
PyramidKV (INT4) & 38\,536 & 20.7763 & 55.78 \\
\bottomrule
\end{tabular}
}
\end{table}

Running the AWQ-Quantized 70B model corroborates these trends. Table~\ref{tab:kv_70b} indicates that full KV quantization alone reduces memory yet slows decode; adding PyramidKV yields a peak usage drop from 40\,690\,MB to 39\,029\,MB while sustaining 57\,tok/s, keeping the end-to-end pipeline within the single-A40 memory budget. Without KV optimization the run exhausts memory before completing 10k-token decodes, underscoring the necessity of pairing sparsification with INT8 cache quantization \citep{hooper2024kvquant,liu2024kivi}.

\section{Coupling Effects in Method Stacking}
\label{sec:coupling}

Budget-aware compression requires a precise view of how modules interact once stacked on a single A40. We observe the strongest positive coupling when ShortGPT depth pruning is applied after AWQ quantization. AWQ's channel protection and scaling calibration depend on activation statistics and error sensitivity under the dense model. If pruning is applied first, it reshapes layer-wise information flow, shifts activation distributions, and reassigns sensitivity, making calibration harder and quantization noise more likely to land in critical directions. Quantizing the full model first yields more stable scaling and protection, while subsequent depth pruning shortens the path over which quantization errors accumulate and propagate. This ordering improves compute efficiency while preserving end-to-end accuracy within a fixed pruning budget (Table~\ref{tab:quant_70b}) \citep{lin2023awq,men2024shortgpt}.
A second positive interaction arises between PyramidKV sparsification and INT8 KV quantization. Sparsification concentrates cache capacity on salient tokens, and quantization then compresses the remaining entries. Together they reduce peak KV memory by over 1.6\,GB relative to the full-KV FP16 baseline while preserving 57 tokens/s on 10k-token prompts (Table~\ref{tab:kv_70b}) \citep{cai2024pyramidkv,hooper2024kvquant}. These synergies directly support the memory and throughput budgets in \Cref{sec:problem} and incur only modest engineering effort, as they reuse existing \texttt{Transformers} hooks and AutoAWQ tooling.

Integration conflicts give rise to negative coupling. Vector-quantization methods such as QTIP and QuIP\# rely on compiler-based optimizations for speed and therefore assume a statically allocated KV cache \citep{tseng2024qtip, tseng2024quipsharp}. In our experiments, enabling these methods disabled dynamic KV caching; although their INT2 compression preserved accuracy and reduced weight size, the achieved throughput fell below a denser AWQ baseline. Moreover, when combined with KV-cache sparsification methods (e.g., SnapKV), the pre-allocation required by static caches prevented any material savings in the actual KV footprint \citep{li2024snapkv}.

Applying unstructured sparsity such as Wanda after AWQ undermines the quantizer's outlier-aware reconstruction of weights. In our early tests, this combination produced unstable decoding on mathematical prompts. The model sometimes emitted nonsensical symbol sequences, and perplexity rose to the thousands rather than the ranges observed for the structured pruning variants (Table~\ref{tab:pruning}) \citep{sun2023wanda, lin2023awq}.

Both conflict types inflate the integration-cost budget: developers must recompile kernels, modify graph boundaries, and in some cases forgo dynamic KV handling altogether. Recognizing these couplings early allows the final pipeline to prioritize compatibility and end-to-end efficiency rather than isolated, per-method benchmark gains.

The contrast between two pipeline explorations underscores these effects. A QTIP-centric pipeline (\emph{INT2 vector quantization + ShortGPT pruning}) achieved extreme compression on screening runs and hinted at deployment on 24\,GB consumer GPUs, yet the combination failed to coexist with PyramidKV, leading to repeated compilation failures and unacceptable integration overhead. In contrast, the AWQ+ShortGPT+PyramidKV configuration sacrifices some compression headroom but remains within memory, throughput, and integration budgets on the A40, making it a practical choice for production. This comparison illustrates the guideline that pipeline-optimal choices differ from method-optimal ones: the strongest standalone quantizer was not selected because of negative coupling, whereas the ultimately adopted trio succeeds precisely because its interactions are benign and budget-aligned.

\section{Unified Pipeline and End-to-End Results}
\label{sec:pipeline}

Guided by the coupling study and the three-dimensional budget, we instantiate a single-GPU pipeline for DeepSeek-R1-Distill-Llama-70B on a single NVIDIA A40.

\paragraph{Backbone compression: quantization and pruning.}
We first apply AutoAWQ (W4A16) to quantize weights, reducing the model to 38\,GB while protecting activation-sensitive channels and matching or exceeding GPTQ on ARC and GPQA (Table~\ref{tab:quant_70b}) \citep{lin2023awq}. Activations remain in FP16 to preserve FlashAttention compatibility and simplify deployment in Transformers.
On top of the AWQ-quantized backbone, we prune 10 of 80 transformer blocks with ShortGPT (12.5\% depth), lowering compute with modest accuracy loss. When additional error is acceptable, we expand the pruning budget using a BI-based, continuity-aware layer reordering that selects low-importance blocks and favors contiguous regions. The resulting checkpoint occupies $\sim$33\,GB, a 75\% reduction relative to the FP16 baseline, while maintaining downstream accuracy within a 5\% margin on broad QA and commonsense tasks.

\paragraph{KV-cache optimization.}
To bound runtime memory at long context, we combine PyramidKV (window size 1024) with INT8 KV-cache quantization \citep{cai2024pyramidkv,hooper2024kvquant}. On 10k-token prompts, this configuration sustains 57.21 tokens/s while peaking at 39\,029\,MB of device memory (Table~\ref{tab:kv_70b}). By contrast, the FP16 baseline exceeds 48\,GB and triggers an out-of-memory failure, underscoring the necessity of cache compression.

\paragraph{Budget summary.}
Composing these stages yields a deployment that fits within our three-dimensional budget from \Cref{sec:problem}. The memory budget is satisfied with peak device memory $B_{\mathrm{mem}} \le 48$\,GB and a 33\,GB model artifact, leaving headroom for prompt buffers and execution context. The throughput budget is met with sustained decode speed above the $B_{\mathrm{thr}} = 57$\,tokens/s target on 10k-token prompts. Finally, all components use open-source implementations with existing integration pathways, so the integration budget $B_{\mathrm{int}}$ is respected: engineering effort is limited to kernel adaptation rather than bespoke compiler work. 

The incremental runs clarify why the final configuration is selected by joint budget satisfaction rather than by a single compression ratio. Dense FP16 and AWQ-only variants exceed the A40 memory budget once the long prompt and KV cache are allocated. AWQ plus depth pruning fits the weight artifact but still leaves long-context decoding exposed to cache growth. Adding PyramidKV and KV quantization shifts the remaining pressure from model storage to runtime state, which is the bottleneck that determines whether the 10k-token setting can run fully on device.

\section{Conclusion}

We presented a unified, budget-aware compression pipeline for deploying 70B-class language models on a single NVIDIA A40 with 10k-token contexts. Our formulation makes three constraints, memory, throughput, and integration, explicit, and uses them to jointly design quantization, pruning, and KV-cache compression. Instantiated with AWQ weight quantization, ShortGPT-style depth pruning, and PyramidKV-based KV compression, the resulting DeepSeek-R1-Distill-Llama-70B deployment fits within a 33\,GB VRAM envelope while maintaining competitive long-context reasoning accuracy.

Beyond combining existing techniques, we show that naive composition is insufficient: unstructured sparsity from Wanda yields limited speedups and destabilizes quantized activations, and aggressively widening or deepening pruning budgets with SliceGPT or ShortGPT leads to substantial accuracy degradation. To address this, we introduce continuity-aware layer reordering and layer-fusion distillation, which exploit architecture-specific BI importance profiles to safely extend depth-only pruning to higher ratios. Together with a modular benchmarking protocol, these components reveal where different methods violate the deployment budget and how their interactions shape end-to-end behavior.

Overall, our study frames compression for large-scale LLMs as a coupled, budget-driven system design problem rather than a collection of independent tricks. The resulting analysis and methodology provide a blueprint for automated pipeline search that jointly optimizes bit-width, pruning ratios, and KV retention under explicit deployment targets.

\section*{Limitations}

Our study is limited to single-GPU inference for a 70B dense checkpoint.
Larger mixture-of-experts systems and consumer GPUs with $\leq$24\,GB VRAM would require additional engineering, and their bottlenecks may differ from the A40 setting studied here.
Our evaluation focuses on English-language QA and reasoning benchmarks.
We do not test multilingual generation, safety-critical domains, or tasks requiring calibrated factuality under distribution shift.
The integration-cost budget is measured in the serving stack used for our experiments.
Engines with native sparse-KV support or specialized vector-quantization kernels could change the trade-off between memory savings and implementation cost.
The proposed recipe is therefore a deployment design for the studied hardware, model family, and 10k-token setting.
It should not be interpreted as a universal compression schedule for all LLM deployments.

\section*{Ethical Considerations}

This paper studies budget-aware compression and systems integration for deploying 70B-class language models on a single GPU with long-context support.
The primary positive impact is improved accessibility.
Reducing memory and compute requirements can lower the cost and energy footprint of deploying capable models, benefiting practitioners with limited hardware and enabling on-device or private deployments where data need not leave a local environment.

At the same time, making large models easier to deploy may increase their availability for harmful uses, including large-scale generation of misleading or spam content.
Our methods optimize inference efficiency but do not mitigate underlying model issues such as bias, privacy leakage, or unsafe content generation.
We therefore recommend standard deployment safeguards, including access control, rate limiting, logging, and content filtering, and compliance with the model license and applicable data governance policies when integrating compressed checkpoints into production systems.

\bibliography{example_paper}

\begin{thebibliography}{34}
\providecommand{\natexlab}[1]{#1}

\bibitem[{Ashkboos et~al.(2024{\natexlab{a}})Ashkboos, Croci, Nascimento,
  Hoefler, and Hensman}]{ashkboos2024slicegpt}
Saleh Ashkboos, Maximilian~L. Croci, Marcelo G.~do Nascimento, Torsten Hoefler,
  and James Hensman. 2024{\natexlab{a}}.
\newblock Slicegpt: Compress large language models by deleting rows and
  columns.
\newblock \emph{arXiv preprint arXiv:2401.15024}.

\bibitem[{Ashkboos et~al.(2024{\natexlab{b}})}]{ashkboos2024quarot}
Saleh Ashkboos and 1 others. 2024{\natexlab{b}}.
\newblock Quarot: Outlier-free 4-bit inference in rotated llms.
\newblock In \emph{NeurIPS}.
\newblock ArXiv:2404.00456.

\bibitem[{Bai et~al.(2023)Bai, Lv, Zhang et~al.}]{bai2023longbench}
Yushi Bai, Xin Lv, Jiajie Zhang, and 1 others. 2023.
\newblock Longbench: A bilingual, multitask benchmark for long context
  understanding.
\newblock \emph{arXiv preprint arXiv:2308.14508}.

\bibitem[{Cai et~al.(2024)}]{cai2024pyramidkv}
Zefan Cai and 1 others. 2024.
\newblock Pyramidkv: Dynamic kv cache compression based on pyramidal
  information funneling.
\newblock \emph{arXiv preprint arXiv:2406.02069}.

\bibitem[{Clark et~al.(2018)Clark, Cowhey, Etzioni, Khot, Sabharwal, Schoenick,
  and Tafjord}]{allenai:arc}
Peter Clark, Isaac Cowhey, Oren Etzioni, Tushar Khot, Ashish Sabharwal, Carissa
  Schoenick, and Oyvind Tafjord. 2018.
\newblock Think you have solved question answering? try arc, the ai2 reasoning
  challenge.
\newblock \emph{arXiv:1803.05457v1}.

\bibitem[{Dao et~al.(2022)Dao, Fu, Ermon, Rudra, and
  R{\'e}}]{dao2022flashattention}
Tri Dao, Daniel~Y. Fu, Stefano Ermon, Atri Rudra, and Christopher R{\'e}. 2022.
\newblock \href {https://arxiv.org/abs/2205.14135} {Flashattention: Fast and
  memory-efficient exact attention with io-awareness}.
\newblock In \emph{NeurIPS}.

\bibitem[{{DeepSeek-AI}(2025)}]{deepseekai2025deepseekr1}
{DeepSeek-AI}. 2025.
\newblock \href {https://arxiv.org/abs/2501.12948} {Deepseek-r1: Incentivizing
  reasoning capability in llms via reinforcement learning}.
\newblock \emph{Preprint}, arXiv:2501.12948.

\bibitem[{Frantar and Alistarh(2023)}]{frantar2023sparsegpt}
Elias Frantar and Dan Alistarh. 2023.
\newblock \href {https://arxiv.org/abs/2301.00774} {Sparsegpt: Massive language
  models can be accurately pruned in one-shot}.
\newblock In \emph{ICML}.

\bibitem[{Frantar et~al.(2022)}]{frantar2022gptq}
Elias Frantar and 1 others. 2022.
\newblock Gptq: Accurate post-training quantization for generative pre-trained
  transformers.
\newblock \emph{arXiv preprint arXiv:2210.17323}.

\bibitem[{Hooper et~al.(2024)Hooper, Kim, Mohammadzadeh, Mahoney, Shao,
  Keutzer, and Gholami}]{hooper2024kvquant}
Coleman Hooper, Sehoon Kim, Hiva Mohammadzadeh, Michael~W Mahoney, Yakun~Sophia
  Shao, Kurt Keutzer, and Amir Gholami. 2024.
\newblock Kvquant: Towards 10 million context length llm inference with kv
  cache quantization.
\newblock \emph{arXiv preprint arXiv:2401.18079}.

\bibitem[{Kwon et~al.(2023)Kwon, Li, Zhuang et~al.}]{kwon2023vllm}
Woosuk Kwon, Zhuohan Li, Siyuan Zhuang, and 1 others. 2023.
\newblock Efficient memory management for large language model serving with
  pagedattention.
\newblock \emph{arXiv preprint arXiv:2309.06180}.

\bibitem[{Li et~al.(2024{\natexlab{a}})Li, Li, Tian, Tang, Xu, Chen, Hu, Dong,
  Li, and Chen}]{li2025surveylargelanguagemodel}
Haoyang Li, Yiming Li, Anxin Tian, Tianhao Tang, Zhanchao Xu, Xuejia Chen,
  Nicole Hu, Wei Dong, Qing Li, and Lei Chen. 2024{\natexlab{a}}.
\newblock \href {https://arxiv.org/abs/2412.19442} {A survey on large language
  model acceleration based on kv cache management}.
\newblock \emph{Preprint}, arXiv:2412.19442.

\bibitem[{Li et~al.(2024{\natexlab{b}})Li, Huang, Yang, Venkitesh, Locatelli,
  Ye, Cai, Lewis, and Chen}]{li2024snapkv}
Yuhong Li, Yingbing Huang, Bowen Yang, Bharat Venkitesh, Acyr Locatelli,
  Hanchen Ye, Tianle Cai, Patrick Lewis, and Deming Chen. 2024{\natexlab{b}}.
\newblock Snapkv: Llm knows what you are looking for before generation.
\newblock \emph{NeurIPS}.
\newblock ArXiv:2404.14469.

\bibitem[{Lin et~al.(2024)}]{lin2024duquant}
Haokun Lin and 1 others. 2024.
\newblock Duquant: Distributing outliers via dual transformation makes stronger
  quantized llms.
\newblock \emph{arXiv preprint arXiv:2406.01721}.

\bibitem[{Lin et~al.(2023)Lin, Xiao et~al.}]{lin2023awq}
Ji~Lin, Guangxuan Xiao, and 1 others. 2023.
\newblock Awq: Activation-aware weight quantization for llm compression and
  acceleration.
\newblock \emph{arXiv preprint arXiv:2306.00978}.

\bibitem[{Liu et~al.(2024)}]{liu2024kivi}
Zirui Liu and 1 others. 2024.
\newblock Kivi: A tuning-free asymmetric 2-bit quantization for kv cache.
\newblock \emph{arXiv preprint arXiv:2402.02750}.

\bibitem[{Men et~al.(2024)Men, Xu, Zhang, Wang, Lin, Lu, Han, and
  Chen}]{men2024shortgpt}
Xin Men, Mingyu Xu, Qingyu Zhang, Bingning Wang, Hongyu Lin, Yaojie Lu, Xianpei
  Han, and Weipeng Chen. 2024.
\newblock Shortgpt: Layers in large language models are more redundant than you
  expect.
\newblock \emph{arXiv preprint arXiv:2403.03853}.

\bibitem[{Merity et~al.(2016)Merity, Xiong, Bradbury, and
  Socher}]{merity2016pointer}
Stephen Merity, Caiming Xiong, James Bradbury, and Richard Socher. 2016.
\newblock \href {https://arxiv.org/abs/1609.07843} {Pointer sentinel mixture
  models}.
\newblock \emph{Preprint}, arXiv:1609.07843.

\bibitem[{Raffel et~al.(2020)Raffel, Shazeer, Roberts, Lee, Narang, Matena,
  Zhou, Li, and Liu}]{raffel2020exploring}
Colin Raffel, Noam Shazeer, Adam Roberts, Katherine Lee, Sharan Narang, Michael
  Matena, Yanqi Zhou, Wei Li, and Peter~J Liu. 2020.
\newblock \href {https://arxiv.org/abs/1910.10683} {Exploring the limits of
  transfer learning with a unified text-to-text transformer}.
\newblock \emph{Journal of machine learning research}, 21(140):1--67.

\bibitem[{Rein et~al.(2024)Rein, Hou, Stickland, Petty, Pang, Dirani, Michael,
  and Bowman}]{rein2024gpqa}
David Rein, Betty~Li Hou, Asa~Cooper Stickland, Jackson Petty, Richard~Yuanzhe
  Pang, Julien Dirani, Julian Michael, and Samuel~R Bowman. 2024.
\newblock \href {https://arxiv.org/abs/2311.12022} {Gpqa: A graduate-level
  google-proof q\&a benchmark}.
\newblock In \emph{First Conference on Language Modeling}.

\bibitem[{Sakaguchi et~al.(2021)Sakaguchi, Bras, Bhagavatula, and
  Choi}]{sakaguchi2021winogrande}
Keisuke Sakaguchi, Ronan~Le Bras, Chandra Bhagavatula, and Yejin Choi. 2021.
\newblock \href {https://arxiv.org/abs/1907.10641} {Winogrande: An adversarial
  winograd schema challenge at scale}.
\newblock \emph{Communications of the ACM}, 64(9):99--106.

\bibitem[{Shao et~al.(2023)}]{shao2023omniquant}
Wenqi Shao and 1 others. 2023.
\newblock Omniquant: Omnidirectionally calibrated quantization for large
  language models.
\newblock \emph{arXiv preprint arXiv:2308.13137}.

\bibitem[{Sun et~al.(2024)Sun, Chang, Bao, Zheng, Zheng, Liu, Dong, Chi, and
  Chen}]{sun2025shadowkvkvcacheshadows}
Hanshi Sun, Li-Wen Chang, Wenlei Bao, Size Zheng, Ningxin Zheng, Xin Liu, Harry
  Dong, Yuejie Chi, and Beidi Chen. 2024.
\newblock \href {https://arxiv.org/abs/2410.21465} {Shadowkv: Kv cache in
  shadows for high-throughput long-context llm inference}.
\newblock \emph{Preprint}, arXiv:2410.21465.

\bibitem[{Sun et~al.(2023)Sun, Liu, Bair, and Kolter}]{sun2023wanda}
Mingjie Sun, Zhuang Liu, Anna Bair, and J.~Zico Kolter. 2023.
\newblock A simple and effective pruning approach for large language models.
\newblock \emph{arXiv preprint arXiv:2306.11695}.

\bibitem[{Tseng et~al.(2024{\natexlab{a}})Tseng, Chee, Sun, Kuleshov, and
  De~Sa}]{tseng2024quipsharp}
Albert Tseng, Jerry Chee, Qingyao Sun, Volodymyr Kuleshov, and Christopher
  De~Sa. 2024{\natexlab{a}}.
\newblock \href {https://arxiv.org/abs/2402.04396} {Quip\#: Even better llm
  quantization with hadamard incoherence and lattice codebooks}.
\newblock In \emph{ICML}.

\bibitem[{Tseng et~al.(2024{\natexlab{b}})}]{tseng2024qtip}
Albert Tseng and 1 others. 2024{\natexlab{b}}.
\newblock Qtip: Quantization with trellises and incoherence processing.
\newblock In \emph{NeurIPS}.
\newblock ArXiv:2406.11235.

\bibitem[{Wan et~al.(2023)Wan, Wang, Liu et~al.}]{wan2024efficientLLM}
Zhongwei Wan, Xin Wang, Che Liu, and 1 others. 2023.
\newblock Efficient large language models: A survey.
\newblock \emph{arXiv preprint arXiv:2312.03863}.

\bibitem[{Wolf et~al.(2020)Wolf, Debut, Sanh, Chaumond, Delangue, Moi, Cistac,
  Rault, Louf, Funtowicz, Davison, Shleifer, von Platen, Ma, Jernite, Plu, Xu,
  Scao, Gugger, Drame, Lhoest, and Rush}]{wolf-etal-2020-transformers}
Thomas Wolf, Lysandre Debut, Victor Sanh, Julien Chaumond, Clement Delangue,
  Anthony Moi, Pierric Cistac, Tim Rault, R{\\'e}mi Louf, Morgan Funtowicz, Joe
  Davison, Sam Shleifer, Patrick von Platen, Clara Ma, Yacine Jernite, Julien
  Plu, Canwen Xu, Teven~Le Scao, Sylvain Gugger, and 3 others. 2020.
\newblock \href {https://www.aclweb.org/anthology/2020.emnlp-demos.6}
  {Transformers: State-of-the-art natural language processing}.
\newblock In \emph{Proceedings of the 2020 Conference on Empirical Methods in
  Natural Language Processing: System Demonstrations}, pages 38--45, Online.
  Association for Computational Linguistics.

\bibitem[{Xiao et~al.(2023{\natexlab{a}})Xiao, Lin, Seznec, Wu, Demouth, and
  Han}]{xiao2022smoothquant}
Guangxuan Xiao, Ji~Lin, Mickael Seznec, Hao Wu, Julien Demouth, and Song Han.
  2023{\natexlab{a}}.
\newblock Smoothquant: Accurate and efficient post-training quantization for
  large language models.
\newblock In \emph{ICML}.
\newblock ArXiv:2211.10438.

\bibitem[{Xiao et~al.(2023{\natexlab{b}})Xiao, Tian, Chen, Han, and
  Lewis}]{xiao2023streamingllm}
Guangxuan Xiao, Yuandong Tian, Beidi Chen, Song Han, and Mike Lewis.
  2023{\natexlab{b}}.
\newblock Efficient streaming language models with attention sinks.
\newblock \emph{arXiv preprint arXiv:2309.17453}.

\bibitem[{Zhang et~al.(2023)Zhang, Sheng, Zhou et~al.}]{zhang2023h2o}
Zhenyu Zhang, Ying Sheng, Tianyi Zhou, and 1 others. 2023.
\newblock H$_2$o: Heavy-hitter oracle for efficient generative inference of
  large language models.
\newblock In \emph{NeurIPS}.
\newblock ArXiv:2306.14048.

\bibitem[{Zheng et~al.(2023)Zheng, Yin, Xie et~al.}]{zheng2023sglang}
Lianmin Zheng, Liangsheng Yin, Zhiqiang Xie, and 1 others. 2023.
\newblock Sglang: Efficient execution of structured language model programs.
\newblock \emph{arXiv preprint arXiv:2312.07104}.

\bibitem[{Zhou et~al.(2024)Zhou, Ning, Hong et~al.}]{zhou2024efficient}
Zixuan Zhou, X~Ning, K~Hong, and 1 others. 2024.
\newblock A survey on efficient inference for large language models.
\newblock \emph{arXiv preprint arXiv:2404.14294}.

\bibitem[{Zhu et~al.(2023)Zhu, Li, Liu, Ma, and
  Wang}]{zhu2024surveymodelcompressionlarge}
Xunyu Zhu, Jian Li, Yong Liu, Can Ma, and Weiping Wang. 2023.
\newblock \href {https://arxiv.org/abs/2308.07633} {A survey on model
  compression for large language models}.
\newblock \emph{Preprint}, arXiv:2308.07633.

\end{thebibliography}

\appendix

\section{Additional Experimental Details}

\subsection{Calibration Corpora and Tasks}
All quantization and pruning calibration follows the same corpora used in Section\,\ref{sec:ptq}. For the 8B screening experiments we sample 128 sequences from WikiText2 and C4 with lengths in \{256, 512\} tokens, mirroring the distributions recommended by the HuggingFace evaluation harness. For DeepSeek-R1-Distill-Llama-70B we reuse 512 prompts drawn from the training mixture provided with the checkpoint, ensuring that the sequences used for calibration are disjoint from the downstream benchmarks (ARC-Challenge/Easy, Winogrande, and GPQA-Diamond). LongBench \citep{bai2023longbench} workloads employ the official validation splits with 11\,427-token contexts to match the memory profiles reported in Table~\ref{tab:kv_70b}.

\subsection{Hardware and Software Configuration}
All measurements are executed on NVIDIA A40 GPUs with 48\,GB of VRAM connected to dual-socket Xeon hosts. We rely on CUDA~12.2, cuBLAS~12.2, and PyTorch~2.3 compiled with compute capability 8.6 support. AutoAWQ provides W4A16 quantization, and the ShortGPT implementation is sourced from the official repository. PyramidKV experiments use the open-source reference implementation with hierarchical cache size set to 1\,024 tokens. Unless otherwise stated, decoding uses a batch size of 1 with greedy sampling; continuous batching is enabled during throughput measurements to match production serving conditions.

\section{Additional Pipeline Validation}
\label{app:pipeline_validation}

This appendix reports additional checks that make the deployment claim easier to audit without changing the main argument.

\paragraph{Incremental end-to-end 70B runs.}
\Cref{tab:pipeline_e2e_70b_full} separates dense, AWQ-only, AWQ plus pruning, and PyramidKV variants under the same 10,240-token A40 setting.
The dense and AWQ-only rows fail once the full prompt and cache are allocated, while the AWQ plus pruning plus PyramidKV rows fit within the device budget.
Adding PyramidKV to the AWQ plus pruning path reduces peak memory from 44.17\,GB to about 42.56\,GB in the unquantized-cache setting; KV quantization reduces memory further but trades off decode speed.

\begin{table*}[t]
\centering
\small
\caption{Full incremental end-to-end results on DeepSeek-R1-Distill-Llama-70B with 10,240 input tokens and 256 generated tokens on one NVIDIA A40. Dense and AWQ-only runs fail because the full prompt and cache exceed the device budget.}
\label{tab:pipeline_e2e_70b_full}
\setlength{\tabcolsep}{4pt}
\resizebox{\textwidth}{!}{%
\begin{tabular}{@{}l c c c r r r r@{}}
\toprule
Configuration & KV quant. & Attention & Artifact size & Peak memory (MB) & Prefill tok/s & Decode tok/s & Total tok/s \\
\midrule
Dense FP16 & -- & -- & 132 GB & OOM & -- & -- & -- \\
AWQ & -- & -- & 38 GB & OOM & -- & -- & -- \\
AWQ + pruning-10 & -- & FA2 & 33 GB & 44,170.07 & 748.23 & 9.49 & 258.20 \\
AWQ + pruning-10 & -- & SDPA & 33 GB & 44,170.07 & 745.45 & 5.07 & 163.47 \\
AWQ + pruning-10 + PyramidKV & -- & FA2 & 33 GB & 42,560.11 & 732.95 & 10.72 & 277.35 \\
AWQ + pruning-10 + PyramidKV & -- & SDPA & 33 GB & 42,560.82 & 731.00 & 10.31 & 270.27 \\
AWQ + pruning-10 + PyramidKV & INT4 & FA2 & 33 GB & 41,880.95 & 519.47 & 8.41 & 209.36 \\
AWQ + pruning-10 + PyramidKV & INT4 & SDPA & 33 GB & 41,878.60 & 426.40 & 8.53 & 194.28 \\
AWQ + pruning-10 + PyramidKV & INT8 & FA2 & 33 GB & 42,174.54 & 539.77 & 8.59 & 215.13 \\
AWQ + pruning-10 + PyramidKV & INT8 & SDPA & 33 GB & 42,169.49 & 561.44 & 8.76 & 221.19 \\
\bottomrule
\end{tabular}}
\end{table*}

\paragraph{Sensitivity of continuity-aware pruning.}
\Cref{tab:pruning_sensitivity} reports nearby continuity settings for the BI-based pruning planner.
The goal is not to claim a universal hyperparameter choice, but to show that the observed improvement over the BI-only ShortGPT baseline is not tied to a single brittle setting.

\begin{table*}[t]
\centering
\small
\caption{Sensitivity of the continuity-aware pruning planner on the AWQ-quantized 70B backbone. The ShortGPT-style rows use BI ranking only. Nearby continuity settings generally improve GPQA-Diamond over the corresponding BI-only baseline, but the best setting remains model dependent.}
\label{tab:pruning_sensitivity}
\setlength{\tabcolsep}{4pt}
\resizebox{\textwidth}{!}{%
\begin{tabular}{@{}l c c c c c c c c@{}}
\toprule
Setting & Pruned layers & $r$ & $\lambda$ & ARC-C & ARC-E & PIQA & WinoGrande & GPQA-D \\
\midrule
Dense AWQ & 0 & -- & -- & $0.5452{\pm}0.0146$ & $0.8068{\pm}0.0081$ & $0.8085{\pm}0.0092$ & $0.7451{\pm}0.0122$ & 0.6313 \\
ShortGPT-style & 15 & -- & -- & $0.4838{\pm}0.0146$ & $0.7534{\pm}0.0088$ & $0.7601{\pm}0.0100$ & $0.7198{\pm}0.0126$ & 0.4848 \\
Continuity-aware & 15 & 1 & 0.10 & $0.4923{\pm}0.0146$ & $0.7588{\pm}0.0088$ & $0.7693{\pm}0.0098$ & $0.7293{\pm}0.0125$ & 0.5303 \\
Continuity-aware & 15 & 1 & 0.05 & $0.5009{\pm}0.0146$ & $0.7500{\pm}0.0089$ & $0.7497{\pm}0.0101$ & $0.7388{\pm}0.0123$ & 0.5303 \\
Continuity-aware & 15 & 1 & 0.20 & $0.5051{\pm}0.0146$ & $0.7622{\pm}0.0087$ & $0.7677{\pm}0.0099$ & $0.7309{\pm}0.0125$ & 0.5152 \\
Continuity-aware & 15 & 3 & 0.10 & $0.4940{\pm}0.0146$ & $0.7601{\pm}0.0088$ & $0.7617{\pm}0.0099$ & $0.7435{\pm}0.0123$ & 0.4747 \\
Continuity-aware & 15 & 3 & 0.20 & $0.4957{\pm}0.0146$ & $0.7551{\pm}0.0088$ & $0.7655{\pm}0.0099$ & $0.7332{\pm}0.0124$ & 0.5202 \\
ShortGPT-style & 20 & -- & -- & $0.4334{\pm}0.0145$ & $0.7012{\pm}0.0094$ & $0.7236{\pm}0.0104$ & $0.7301{\pm}0.0125$ & 0.2879 \\
Continuity-aware & 20 & 1 & 0.10 & $0.4565{\pm}0.0146$ & $0.7247{\pm}0.0092$ & $0.7454{\pm}0.0102$ & $0.7261{\pm}0.0125$ & 0.3838 \\
Continuity-aware & 20 & 1 & 0.05 & $0.4701{\pm}0.0146$ & $0.7184{\pm}0.0092$ & $0.7410{\pm}0.0102$ & $0.7206{\pm}0.0126$ & 0.3485 \\
Continuity-aware & 20 & 3 & 0.20 & $0.4113{\pm}0.0144$ & $0.6481{\pm}0.0098$ & $0.7100{\pm}0.0106$ & $0.7230{\pm}0.0126$ & 0.3131 \\
\bottomrule
\end{tabular}}
\end{table*}

\paragraph{Integration-budget checklist.}
\Cref{tab:integration_checklist} expands the qualitative integration budget into concrete implementation checks: kernel availability, dynamic KV-cache compatibility, compiler assumptions, and the batch/context regime.
The target setting is fully on-device 10k-token inference on one A40, where batch size 1 is already memory constrained.

\begin{table*}[t]
\centering
\small
\caption{Integration-budget checklist used in place of a single subjective integration score. A method is considered compatible with the target pipeline only if it can be measured under the same fully on-device 10k-token A40 setting.}
\label{tab:integration_checklist}
\setlength{\tabcolsep}{5pt}
\begin{tabular}{@{}p{0.18\linewidth} p{0.19\linewidth} p{0.23\linewidth} p{0.32\linewidth}@{}}
\toprule
Checklist item & Low-friction case & High-friction case & How it affects this study \\
\midrule
Kernel path & Uses existing attention and GEMM kernels & Requires custom low-bit or sparse kernels & AWQ and depth pruning reuse existing kernels; SliceGPT and vector quantizers require more kernel-specific work. \\
KV-cache layout & Works with dynamic cache retention & Assumes static preallocated cache & PyramidKV needs dynamic cache management; this conflicts with the fast path of some vector-quantized deployments. \\
Compiler assumptions & Stable under eager or standard compiled execution & Requires fixed shapes or graph rewrites & Static-shape assumptions can disable dynamic KV sparsification or force recompilation. \\
Batching target & Valid at batch size 1 and 10k prompts & Valid mainly for short-context high-batch serving & Our target is long-context local inference, so high-batch serving throughput is not the primary metric. \\
\bottomrule
\end{tabular}
\end{table*}

\paragraph{Scope of serving-system comparisons.}
Serving systems optimize related but distinct operating points, so we do not claim a general advantage over them.
vLLM with AWQ is a strong paged-KV serving baseline, but in our fixed 10,240-token A40 target the AWQ-only path remains out of memory without the PyramidKV stage.
llama.cpp and ExLlamaV2 are important low-bit inference paths, but CPU/GPU offload and specialized low-bit kernels change the budget through host memory, PCIe bandwidth, and dynamic sparse-KV support.

\section{Reproducibility Notes}

We fix pseudo-random seeds to 42 across PyTorch, NumPy, and the evaluation harness. Measurements are taken after discarding the first 32 generated tokens to avoid cold-start effects, and each configuration is profiled for at least three full decoding passes to ensure stable NVML readings. The preprocessing scripts that normalize calibration corpora and the YAML files that define each pipeline stage are available upon request; they contain no proprietary assets and depend only on public model checkpoints. Hyperparameters for PyramidKV (window size 512, retention ratio 0.25) and KV quantization (symmetric INT8 with per-head scaling) are shared with the code release of the respective papers to facilitate cross-study comparisons.

We also log failed allocation states rather than only successful generations. OOM rows in \Cref{tab:pipeline_e2e_70b_full} are counted only after the model, prompt tensors, and cache allocation reach the same sequence-length target used by the successful rows. This prevents a partially loaded model from being treated as a comparable deployment point.

The reported pipeline uses public checkpoints and public implementations. We do not redistribute third-party checkpoints, datasets, or derived model weights; artifact access follows the original public release pages and their stated terms. When a method requires a specialized runtime or a fixed-shape compiler path, we record that requirement as part of the integration budget instead of treating it as a timing-only detail. This convention keeps memory, speed, and implementation assumptions tied to the same deployment setting.

For throughput, we keep prefill and decode speeds separate in the raw logs before reporting total tokens per second. This separation matters because cache-compression methods often shift cost between the two phases. The final tables therefore report enough detail to distinguish a method that only accelerates prefill from one that improves the sustained decoding regime targeted by this paper.

\end{document}